\documentclass[conference]{IEEEtran}
\IEEEoverridecommandlockouts

\usepackage{amsmath,amssymb,amsfonts}
\usepackage{graphicx}
\usepackage{textcomp}
\usepackage{xcolor}

\usepackage{booktabs}
\usepackage{multirow}
\usepackage{siunitx}
\usepackage{threeparttable}
\usepackage{colortbl}

\graphicspath{{Image/}}
\usepackage{subcaption}
\usepackage{placeins}

\usepackage{algorithm}
\usepackage{algorithmic}

\usepackage[hidelinks]{hyperref}

\allowdisplaybreaks

\begin{document}

\title{Prune-Quantize-Distill: An Ordered Pipeline for Efficient Neural Network Compression}

\author{
\IEEEauthorblockN{1\textsuperscript{st} Longsheng Zhou}
\IEEEauthorblockA{\textit{Institute of Advanced Technology} \\
\textit{University of Science and Technology of China} \\
Hefei, China \\
zlongsheng@mail.ustc.edu.cn}
\and
\IEEEauthorblockN{2\textsuperscript{nd} Yu Shen\thanks{\textsuperscript{*}Corresponding author: shenyu@ustc.edu.cn}}
\IEEEauthorblockA{\textit{Supercomputing Center} \\
\textit{University of Science and Technology of China} \\
Hefei, China \\
shenyu@ustc.edu.cn}
}

\maketitle

\bstctlcite{IEEEexample:BSTcontrol}

\begin{abstract}
Modern deployment often requires trading accuracy for efficiency under tight CPU and memory constraints, yet common compression proxies such as parameter count or FLOPs do not reliably predict wall-clock inference time. In particular, unstructured sparsity can reduce model storage while failing to accelerate (and sometimes slightly slowing down) standard CPU execution due to irregular memory access and sparse-kernel overhead. Motivated by this gap between \emph{compression} and \emph{acceleration}, we study a practical, ordered pipeline that targets \emph{measured} latency by combining three widely used techniques: unstructured pruning, INT8 quantization-aware training (QAT), and knowledge distillation (KD). Empirically, INT8 QAT provides the dominant runtime benefit, while pruning mainly acts as a capacity-reduction pre-conditioner that improves the robustness of subsequent low-precision optimization; KD, applied last, recovers accuracy within the already constrained sparse INT8 regime without changing the deployment form. We evaluate on CIFAR-10/100 using three backbones (ResNet-18, WRN-28-10, and VGG-16-BN). Across all settings, the ordered pipeline achieves a stronger accuracy--size--latency frontier than any single technique alone, reaching 0.99--1.42\,ms CPU latency with competitive accuracy and compact checkpoints. Controlled ordering ablations with a fixed 20/40/40 epoch allocation further confirm that stage order is consequential, with the proposed ordering generally performing best among the tested permutations. Overall, our results provide a simple guideline for edge deployment: evaluate compression choices in the joint accuracy--size--latency space using measured runtime, rather than proxy metrics alone.
\end{abstract}


\section{Introduction}
\label{sec:introduction}

Deep neural networks (DNNs) have achieved strong performance in vision, language, and multimodal tasks, but modern models are often over-parameterized and expensive to run. This becomes a real obstacle on resource-constrained platforms such as mobile devices, embedded systems, and edge accelerators, where memory footprint, power budget, and inference latency are all limited. In these settings, simply shrinking the backbone (fewer layers or channels) can quickly lead to noticeable accuracy degradation.

Model compression provides a more controlled way to trade a small amount of accuracy for substantial efficiency gains~\cite{zongshu1}. Recent surveys summarize a wide range of compression techniques and emphasize that deployment constraints (hardware backend, operator support, and runtime overhead) often dominate practical outcomes~\cite{zongshu2}. A comparative study by Marin\`o~\textit{et al.} further suggests that different compression families typically improve different parts of the accuracy--efficiency space and that hybrid strategies are often preferable to single techniques~\cite{zongshu3}.

In this work, we focus on three widely used building blocks: pruning, quantization, and knowledge distillation (KD).
Pruning removes redundant parameters and can substantially reduce effective capacity with limited accuracy loss under appropriate retraining~\cite{Prune2}, while its practical speed benefits depend on sparsity structure and hardware support~\cite{duobuzhou1}.
Quantization reduces precision and storage cost, enabling efficient inference on standard integer backends (e.g., INT8)~\cite{zhou2017incremental}.
KD transfers knowledge from a high-capacity teacher to a constrained student and is often used to recover accuracy after compression~\cite{Distill1,Distill2}.
We deliberately restrict ourselves to these standard components, aiming for a minimal and reproducible pipeline rather than introducing specialized operators or elaborate training tricks.

Despite steady progress, combining these tools in a reliable way is still tricky.
Many hybrid approaches are tailored to particular architectures or datasets, or rely on tightly coupled joint objectives and careful hyperparameter tuning~\cite{duobuzhou4}.
Other hybrid ideas target different model families with different deployment bottlenecks, which may not directly carry over to standard CNN inference~\cite{2.3.2}.
Meanwhile, practitioners often prefer simple recipes built from standard components, because they are easier to reproduce and integrate into existing toolchains~\cite{liang2021pruning}.
This motivates a concrete question: can we design a minimal hybrid pipeline that consistently improves the joint accuracy--size--latency trade-off on mainstream convolutional backbones under measured CPU runtime, without specialized sparse kernels or intricate training tricks?

We propose a fixed three-stage recipe: \emph{global unstructured pruning} $\rightarrow$ \emph{INT8 quantization-aware training (QAT)} $\rightarrow$ \emph{knowledge distillation}, where the ordering is part of the method and targets a consistent deployable form (sparse INT8).
Unstructured pruning alone often yields limited wall-clock benefits on general-purpose CPUs \emph{without specialized sparse kernels}, but it reduces the active weight set and stabilizes subsequent low-precision optimization.
INT8 QAT contributes most of the latency reduction on standard backends, and KD is applied last to recover accuracy after the model has entered the constrained sparse INT8 regime.
Overall, the stages play complementary roles, and we validate the importance of ordering via a controlled stage permutation study.

We evaluate the pipeline on three backbone--dataset pairs: ResNet-18 on CIFAR-10, WRN-28-10 on CIFAR-100, and VGG-16-BN on CIFAR-10.
We compare against pruning-only, quantization-only, and KD-only baselines, and analyze the joint accuracy--size--latency space using both scalar metrics and multi-dimensional visualizations.
In addition, to align with common co-compression benchmarks in prior work, we report a literature-aligned comparison on ResNet-20/CIFAR-10 using relative BOPs.

\subsection*{Our main contributions are summarized as follows:}
\begin{itemize}
    \item \textbf{A minimal ordered co-compression recipe with role separation.}
    We propose a simple three-stage pipeline---global unstructured pruning $\rightarrow$ INT8 QAT $\rightarrow$ KD---built from standard components and evaluated at a consistent deployable endpoint (sparse INT8), avoiding reliance on specialized sparse kernels.

    \item \textbf{Controlled evidence that stage ordering is consequential.}
    Keeping the same ingredients, the same training budget, and the same deployable form, we perform a controlled stage-order ablation by permuting the three stages, showing that ordering alone can lead to clear accuracy differences while leaving latency within a narrow range.

    \item \textbf{Deployment-driven evaluation with consistent gains across settings.}
    We evaluate compression choices in the joint accuracy--size--latency space using measured CPU runtime, demonstrating that proxy metrics (e.g., parameter/FLOP reductions) may not reflect wall-clock behavior on standard backends.
    Across three CNN backbones and two datasets, the proposed ordering yields consistently favorable trade-offs over single-stage baselines; we additionally report a literature-aligned comparison using relative BOPs on ResNet-20/CIFAR-10 to corroborate the conclusions under a common proxy metric.
\end{itemize}


\begin{figure*}[!t]
    \centering
    \includegraphics[width=0.95\textwidth]{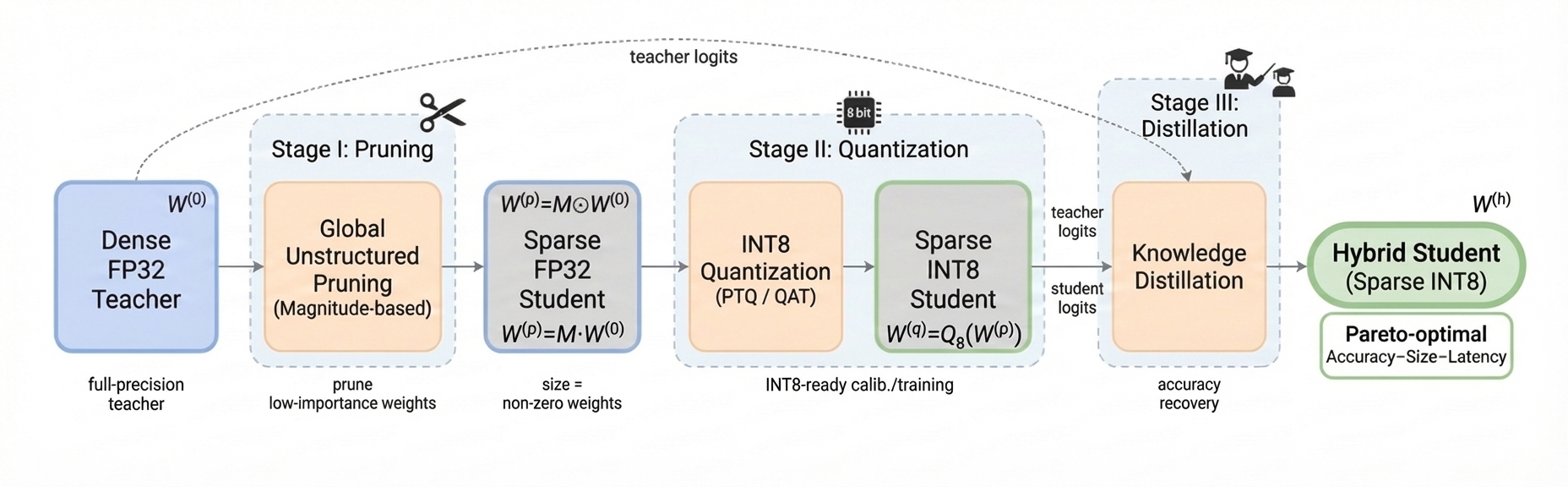}
    \caption{Overview of the proposed hybrid compression pipeline. A dense FP32 teacher with weights $\mathbf{W}^{(0)}$ is first pruned into a sparse FP32 student ($\mathbf{W}^{(p)} = \mathbf{M} \odot \mathbf{W}^{(0)}$), then optimized under INT8 quantization (PTQ/QAT) to obtain a sparse INT8 student $\mathbf{W}^{(q)} = Q_8(\mathbf{W}^{(p)})$, and finally refined via knowledge distillation to produce the hybrid student $\mathbf{W}^{(h)}$ in the compressed space.}
    \label{fig:pipeline}
\end{figure*}

\section{Related Work} 
\label{sec:related}

We review the three standard ingredients used in our pipeline---pruning, quantization, and knowledge distillation---and then briefly discuss hybrid, multi-stage recipes. \subsection{Pruning} Pruning removes parameters (or structures) that contribute little to the final prediction. Early magnitude-based pruning with retraining showed that CNNs can often be made much smaller with only minor accuracy loss. Later work studied pruning together with quantization, including structured or coordinated pruning--quantization schemes~\cite{jianzhinew1}. In our setting, we care more about preserving accuracy on compact backbones than about relying on sparse kernels for speed. We therefore use global unstructured pruning mainly as an accuracy-friendly way to reduce the number of active weights before quantization, rather than treating pruning as the main source of wall-clock acceleration~\cite{jianzhinew2}.Recent analyses further suggest that unstructured sparsity mainly reduces storage and may not translate to latency gains on general-purpose CPUs unless supported by specialized kernels, which motivates using pruning primarily as a capacity regularizer in our pipeline. \subsection{Quantization} Quantization speeds up inference by using low-precision integers for weights and activations. Early studies showed that quantized approximations can reduce test-time cost while keeping accuracy reasonably close to full precision~\cite{qatnew1}. Incremental schemes then pushed pretrained CNNs to low precision more gently, reducing the drop in accuracy. Deep Compression combined trained quantization with coding to further shrink the stored model~\cite{qatnew3}. A common issue is that aggressive post-training discretization can be brittle. For this reason, we use INT8 quantization-aware training (QAT) and optimize the model directly under fake-quant constraints, starting from a pruned initialization.Empirically, a pruned initialization can also ease low-precision optimization by lowering the effective noise accumulation, making INT8 training more stable than quantizing a dense model directly. \subsection{Knowledge Distillation} Knowledge distillation (KD) trains a student to match a stronger teacher, commonly through a KL loss on softened outputs together with cross-entropy. KD is often used to boost small or compressed models, including pruned networks, and can recover part of the accuracy lost during compression~\cite{zhengliunew2}. KD itself does not change size or latency, but it is a practical tool for “repairing” a student after pruning/quantization. We therefore apply KD at the end, inside the sparse INT8 space, to regain accuracy without changing the deployment cost.This “post-compression refinement” view is especially useful when quantization introduces functional shifts that are hard to correct by fine-tuning alone. \subsection{Hybrid and Multi-Stage Compression Frameworks} Many recent works combine multiple compression steps rather than relying on a single technique. Beyond the choice of ingredients (pruning/quantization/distillation), existing methods also differ substantially in their deployment assumptions and optimization paradigms, e.g., fixed-bit versus mixed-precision quantization, structured versus unstructured pruning, and staged pipelines versus constrained (white-box) formulations or coupled update rules~\cite{zuhenew3}. To make these design choices explicit, we summarize representative paradigms in Table~\ref{tab:design_philosophy_compact}. In contrast to tightly coupled co-optimization approaches that often rely on joint objectives and careful tuning, we adopt a minimal and strictly ordered recipe: global unstructured pruning $\rightarrow$ INT8 quantization-aware training (QAT) $\rightarrow$ knowledge distillation in the compressed space. This design favors fixed-bit INT8 deployment and uses KD as a lightweight accuracy recovery step without changing deployment cost. Overall, our framework aligns with the broader trend of staged compression, but emphasizes a clearer separation of the roles of pruning, quantization, and distillation, thereby characterizing how different combinations and ordering choices affect the final accuracy--efficiency trade-off.

\begin{table}[t]
\centering
\caption{Design philosophy comparison (compact).}
\label{tab:design_philosophy_compact}
\setlength{\tabcolsep}{3.2pt} 
\renewcommand{\arraystretch}{1.08}
\footnotesize
\begin{tabular}{lcccc}
\hline
\textbf{Method} & \textbf{INT8} & \textbf{Prune} & \textbf{Opt.} & \textbf{KD} \\
\hline
\textbf{Ours (P50$\to$INT8$\to$KD)} & \textbf{$\checkmark$} & Unstr. & Pipeline & \textbf{$\checkmark$} \\
GETA & $\triangle$ & \textbf{Str.} & Joint & $\times$ \\
SQL  & $\times$ & Unstr. & Constr. & $\times$ \\
QST  & $\times$ & Unstr. & Coupled & $\times$ \\
ANNC & $\triangle$ & Unstr. & Constr. & $\times$ \\
\hline
\end{tabular}

\vspace{2pt}
\scriptsize \textit{Notes:} INT8: $\checkmark$ fixed single-precision INT8; $\times$ primarily mixed-precision / bit-allocation; $\triangle$ configuration-dependent. Str./Unstr.: structured/unstructured pruning. Opt.: Constr.=constrained (white-box) optimization.
\end{table}

\section{Method}
\label{sec:method}

\subsection{Problem Formulation and Hybrid View}
\label{subsec:method_overview}

We use three metrics throughout the paper: accuracy, effective model size, and inference latency. Effective size is measured by the number of non-zero parameters; we also report the serialized checkpoint footprint (MB) as a practical storage proxy (Tables~\ref{tab:tradeoff_multisetting}--\ref{tab:tradeoff_resnet18}). Fig.~\ref{fig:pipeline} summarizes the proposed three-stage procedure.

Let $f(\mathbf{x};\mathbf{W})$ denote a classifier trained on $\{(\mathbf{x}_i,y_i)\}_{i=1}^N$. Standard supervised learning minimizes
\begin{equation}
\min_{\mathbf{W}} \ \mathcal{L}(\mathbf{W})
= \frac{1}{N}\sum_{i=1}^{N} \ell\big(f(\mathbf{x}_i;\mathbf{W}),y_i\big),
\end{equation}
where $\ell(\cdot)$ is the cross-entropy loss.

Deployment imposes sparsity, INT8-deployability, and latency constraints:
\begin{equation}
\min_{\mathbf{W}} \ \mathcal{L}_{\text{task}}(\mathbf{W})
\quad\text{s.t.}\quad 
\mathbf{W}\in \mathcal{S}_\rho,\ \mathbf{W}\in \mathcal{Q}_8,\ 
\text{Lat}(\mathbf{W})\le \tau ,
\label{eq:feasible_constraint}
\end{equation}
where $\mathcal{S}_\rho$ is the set of models with target sparsity $\rho$, $\mathcal{Q}_8$ denotes INT8-deployable models, and $\tau$ is a latency budget.

Directly solving the above constrained optimization problem is inconvenient because sparsity is non-smooth and INT8 quantization is discrete.
A useful way to summarize the interaction between the three stages is the coupled objective
\begin{equation}
\mathbf{W}^\star \approx
\arg\min_{\mathbf{W}}
\mathcal{L}_{\text{task}}\big(Q_8(\mathbf{M}\odot\mathbf{W})\big)
+\lambda_s\|\mathbf{M}\odot\mathbf{W}\|_0
+\lambda_d\,\mathcal{L}_{KD},
\label{eq:hybrid_objective}
\end{equation}
where $\mathbf{M}\in\{0,1\}^{|\mathbf{W}|}$ is a pruning mask and $\mathcal{L}_{KD}$ is defined in Sec.~\ref{subsec:kd}.

In practice, we approximate this coupled objective with a fixed three-stage pipeline:
\begin{equation}
\mathbf{W}^{(0)} 
\;\xrightarrow{\text{Pruning}}\;
\mathbf{W}^{(p)}
\;\xrightarrow{\text{QAT}}\;
\mathbf{W}^{(q)}
\;\xrightarrow{\text{KD}}\;
\mathbf{W}^{(h)} ,
\label{eq:pipeline}
\end{equation}
which can be interpreted as moving the model into progressively smaller feasible sets:
\begin{equation}
\mathbb{R}^{|\mathbf{W}|} \supset \mathcal{S}_\rho \supset \mathcal{S}_\rho\cap\mathcal{Q}_8.
\label{eq:feasible_nesting}
\end{equation}
The ordering matters because the final stage operates inside the deployable sparse INT8 space, rather than correcting errors in an unconstrained FP32 model.

\subsection{Stage I: Global Unstructured Magnitude Pruning}
\label{subsec:pruning}

We employ global unstructured magnitude pruning. With a binary mask $\mathbf{M}$,
\begin{equation}
\mathbf{W}^{(p)} = \mathbf{M}\odot \mathbf{W}^{(0)}, \qquad
\|\mathbf{W}^{(p)}\|_0 = \sum_j M_j ,
\label{eq:sparse_weight}
\end{equation}
and the mask keeps the top $(1-\rho)$ weights by magnitude:
\begin{equation}
M_j = \mathbb{I}\big(|W^{(0)}_j|\ge \gamma_\rho\big), \quad
\frac{\|\mathbf{W}^{(p)}\|_0}{\|\mathbf{W}^{(0)}\|_0}=1-\rho,
\label{eq:global_prune}
\end{equation}
where $\gamma_\rho$ is the threshold for sparsity $\rho$.
A common first-order argument relates pruning $W_j$ to the loss change:
\begin{equation}
\Delta \mathcal{L}\approx 
\left|\frac{\partial \mathcal{L}}{\partial W_j} W_j\right|.
\label{eq:taylor_prune}
\end{equation}

\noindent\textbf{Why unstructured pruning?}
Structured pruning can give direct speedups, but it often degrades accuracy on compact backbones. Unstructured pruning typically preserves accuracy better, even though it may not reduce CPU latency due to irregular memory access. In this paper it is used mainly to reduce the active weight set and to make the later INT8 optimization less noisy.

\subsection{Stage II: INT8 Quantization-Aware Training (QAT)}
\label{subsec:qat}

On top of $\mathbf{W}^{(p)}$, we perform INT8 QAT using uniform affine quantization:
\begin{equation}
Q_{8}(x)=\mathrm{clip}\!\left(\left\lfloor\frac{x}{s}\right\rceil+z,\;0,\;255\right),
\qquad
\hat{x}=s\,(Q_{8}(x)-z),
\label{eq:int8_quant}
\end{equation}
where $s$ is the scale and $z$ is the zero-point. QAT minimizes the task loss under fake-quant constraints,
\begin{equation}
\min_{\mathbf{W}} \;
\mathcal{L}_{\text{task}}
\Big(f\big(\mathbf{x}; Q_{8}(\mathbf{M}\odot\mathbf{W})\big)\Big),
\label{eq:qat_objective}
\end{equation}
and uses the straight-through estimator (STE),
\begin{equation}
\frac{\partial Q_8(x)}{\partial x}\approx 1.
\label{eq:ste}
\end{equation}

\paragraph{Quantization on top of pruning}
Pruning reduces the number of active weights,
\begin{equation}
\|\mathbf{W}^{(p)}\|_0=(1-\rho)\|\mathbf{W}^{(0)}\|_0,
\label{eq:sparsity_ratio}
\end{equation}
and QAT quantizes the remaining weights to INT8,
\begin{equation}
\mathbf{W}^{(q)} = Q_8(\mathbf{W}^{(p)}).
\label{eq:quant_after_prune}
\end{equation}
Ignoring sparse storage overhead, this yields an approximate multiplicative size reduction:
\begin{equation}
\frac{\text{Size}(\mathbf{W}^{(q)})}{\text{Size}(\mathbf{W}^{(0)})}
\approx
(1-\rho)\cdot \frac{8}{32},
\qquad
\text{Compr.}\approx \frac{32}{8}\cdot\frac{1}{1-\rho}.
\label{eq:mul_compression}
\end{equation}

\paragraph{Why pruning can stabilize QAT}
Pruning may also reduce the aggregate effect of quantization perturbations by shrinking the active set. Under a standard uniform error model $Q_8(w)=w+\epsilon$ with $\epsilon\sim\mathcal{U}(-\Delta/2,\Delta/2)$,
\begin{equation}
\mathbb{E}\|\boldsymbol{\epsilon}\|_2^2
\le \frac{\Delta^2}{12}\cdot \|\mathbf{W}^{(p)}\|_0.
\label{eq:quant_noise_bound}
\end{equation}
This provides an intuitive motivation (not a guarantee): fewer active weights can reduce the accumulated perturbation during INT8 optimization, which aligns with our empirical ordering ablation.

\paragraph{PTQ vs.\ QAT}
We mainly report QAT since it is typically more robust than PTQ under aggressive compression. PTQ remains a training-free deployment option on the same compressed backbone.

\subsection{Stage III: Knowledge Distillation for Accuracy Recovery}
\label{subsec:kd}

Pruning and quantization introduce a functional deviation from the dense teacher:
\begin{equation}
\Delta f(\mathbf{x})
=
f(\mathbf{x};\mathbf{W}^{(q)})
-
f(\mathbf{x};\mathbf{W}^{(0)}).
\label{eq:function_shift}
\end{equation}
KD uses the dense teacher to guide the fake-quantized student. With logits $\mathbf{z}_t$ and $\mathbf{z}_s$,
\begin{equation}
\mathcal{L}_{KD}
= \alpha\,\mathcal{L}_{CE}
+ (1-\alpha)T^2 \, \mathrm{KL}\!\left(
\sigma(\mathbf{z}_t/T)\,\|\,\sigma(\mathbf{z}_s/T)
\right),
\label{eq:kd_loss}
\end{equation}
where $T$ is temperature and $\alpha$ balances CE and KD.

KD refines the student within the sparse INT8 feasible set:
\begin{equation}
\mathbf{W}^{(h)}
=
\arg\min_{\mathbf{W}\in\mathcal{S}_\rho\cap\mathcal{Q}_8}
\mathbb{E}_{\mathbf{x}}
\|f(\mathbf{x};\mathbf{W})-f(\mathbf{x};\mathbf{W}^{(0)})\|_2^2,
\label{eq:kd_feasible}
\end{equation}
and does not change effective size or latency:
\begin{equation}
\|\mathbf{W}^{(h)}\|_0=\|\mathbf{W}^{(q)}\|_0,\quad
\text{Lat}(\mathbf{W}^{(h)})\approx\text{Lat}(\mathbf{W}^{(q)}).
\label{eq:kd_no_cost}
\end{equation}

In practice, we distill from the original dense FP32 teacher $\mathbf{W}^{(0)}$ to avoid propagating quantization artifacts from a low-precision teacher. We found logit-based distillation to be sufficient for compact CNN backbones, since it directly targets the decision boundary shift induced by pruning and INT8 constraints (Eq.~\ref{eq:function_shift}). KD is applied with the same fake-quant modules enabled as in QAT, so the student is optimized in the exact deployable space rather than in an unconstrained surrogate. This differs from distilling before quantization, where improvements may partially vanish after discretization. Unless otherwise stated, we keep the KD setup lightweight (teacher fixed, student initialized from $\mathbf{W}^{(q)}$) and tune only $(\alpha, T)$.
We also note that applying KD earlier in the pipeline, such as before pruning or before quantization, leads to less consistent gains in our experiments. When KD is performed on a dense or FP32 model, the distilled knowledge may not be preserved after subsequent pruning or discretization. By contrast, performing KD as the final stage allows the student to adapt its predictions to the combined effects of sparsity and INT8 quantization. From an optimization perspective, KD acts as a local refinement step that reshapes the loss landscape within the constrained feasible region. This makes it particularly effective for recovering accuracy without reintroducing additional parameters or increasing inference cost.

\subsection{Complementarity and Ordering}
\label{subsec:ordering}

The stages play complementary roles: pruning shrinks the active set and stabilizes subsequent INT8 optimization, QAT provides the main speedup, and KD recovers accuracy after the model has entered the sparse INT8 regime. Importantly, the \emph{order} of these stages matters: under a controlled ordering ablation where we fix the same components and stage budgets (20/40/40 epochs) and only permute the stage order, the default ordering (Prune$\rightarrow$QAT$\rightarrow$KD) consistently achieves the best accuracy across backbones (Table~\ref{tab:tradeoff_multisetting}). Overall, the resulting sparse INT8 model occupies a strong region in the joint accuracy--size--latency space.

\begin{figure*}[!t]
    \centering
    \includegraphics[width=0.95\textwidth]{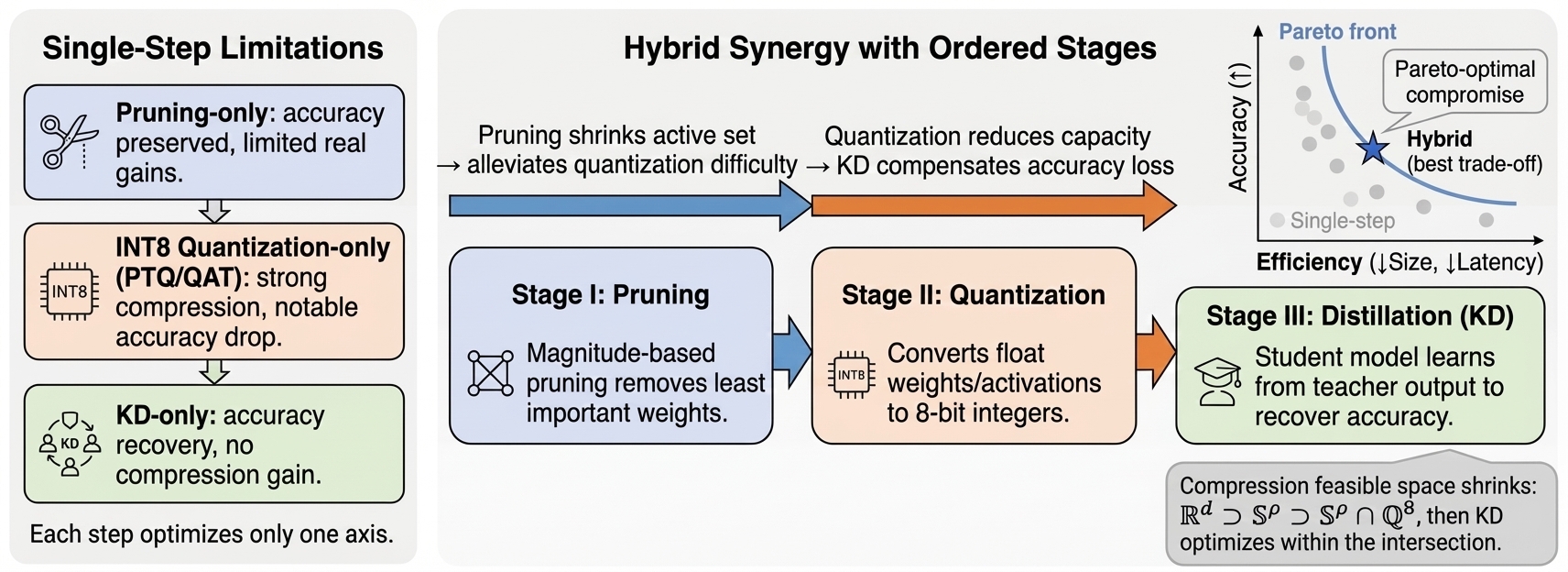}
    \caption{Synergy of the ordered hybrid method. 
    Left: limitations of single-step pruning-only, quantization-only, and KD-only strategies, each optimizing mainly one axis (accuracy or efficiency).
    Right: our three-stage ordering, where pruning shrinks the active set, quantization converts sparsity into real efficiency gains, and KD recovers accuracy in the compressed space, moving the model toward the Pareto front in the accuracy--efficiency plane.}
    \label{fig:synergy}
\end{figure*}

\section{Experiments}
\label{sec:experiments}

\subsection{Experimental Setup}
\label{subsec:exp_setup}

We conduct our main experiments on three backbone--dataset pairs: ResNet-18 on CIFAR-10, WideResNet-28-10 (WRN-28-10) on CIFAR-100, and VGG-16-BN on CIFAR-10.
In addition, to align with common co-compression benchmarks in prior work, we report a literature-aligned comparison on ResNet-20/CIFAR-10.
For each backbone, we first train a dense FP32 network as the \emph{teacher} and as the reference point for reporting compression and speedup.
Unless otherwise stated, the FP32 baseline is optimized for 100 epochs using SGD with cosine learning-rate decay.

\paragraph{Budgets and reporting protocols}
We report results under three complementary protocols.
\textbf{(i) Fully trained (Table~\ref{tab:tradeoff_multisetting}):} all methods are compared under a fixed total budget of 100 epochs.
For the multi-stage pipeline, we allocate epochs across stages while keeping the total budget fixed (20/40/40 for Prune/QAT/KD).
\textbf{(ii) Diagnostic snapshot (Table~\ref{tab:tradeoff_resnet18}):} an early-training snapshot used to diagnose the roles and interactions of individual stages in a multi-stage pipeline, rather than to claim final performance.
\textbf{FT Epochs} denotes the number of fine-tuning epochs performed \emph{on top of the baseline checkpoint}; for multi-stage pipelines, fine-tuning epochs are counted and summed across stages.
\textbf{(iii) Literature-aligned benchmark (Table~\ref{tab:cifar10_rn20_300e_bops}):} ResNet-20 on CIFAR-10, reported with accuracy and relative BOPs to match common co-compression settings.

\emph{Reading guide:} cross-backbone conclusions (including ordering ablations) are drawn from the fully trained protocol (Table~\ref{tab:tradeoff_multisetting});single-backbone diagnostic discussions use the snapshot protocol (Table~\ref{tab:tradeoff_resnet18});  and the literature-aligned benchmark (Table~\ref{tab:cifar10_rn20_300e_bops}) is used only for comparison under standard co-compression reporting.

Starting from the same FP32 teacher initialization, we derive four variants:
(i) pruning-only, (ii) QAT-only, (iii) KD-only, and (iv) the ordered hybrid pipeline (Prune$\rightarrow$QAT$\rightarrow$KD).
We further include an \textbf{ordering ablation} by permuting the three stages while fixing the same components and the same 20/40/40 stage budgets (Table~\ref{tab:tradeoff_multisetting}).

\paragraph{Latency measurement}
Inference latency is measured end-to-end on a single Intel Xeon (Skylake) CPU server under a unified PyTorch setup.
All INT8 models are evaluated with the \texttt{fbgemm} backend, and FP32 models run on the same CPU under identical threading settings.
We fix PyTorch to use 10 CPU threads, switch models to evaluation mode, and time forward passes with \texttt{time.perf\_counter()} inside \texttt{torch.inference\_mode()}.
Each latency number is averaged over 100 repeated runs after 10 warm-up runs; we reuse the same test batch across repeats to reduce batch-content variance.

Effective model size is measured by the number of non-zero parameters and the serialized checkpoint footprint (in MB), which we denote as \emph{Size (MB)} in the tables.

\begin{table*}[t]
\centering
\caption{Main results (fully trained) across backbones and datasets with controlled stage-order ablation.}
\label{tab:tradeoff_multisetting}

\footnotesize
\setlength{\tabcolsep}{3.0pt}
\renewcommand{\arraystretch}{1.10}

\resizebox{\textwidth}{!}{
\begin{tabular}{lcccc cccc cccc}
\toprule
\multirow{2}{*}{\textbf{Method}}
& \multicolumn{4}{c}{\textbf{ResNet-18 + CIFAR-10}}
& \multicolumn{4}{c}{\textbf{WRN-28-10 + CIFAR-100}}
& \multicolumn{4}{c}{\textbf{VGG-16-BN + CIFAR-10}} \\
\cmidrule(lr){2-5}\cmidrule(lr){6-9}\cmidrule(lr){10-13}
& Acc. (\%) $\uparrow$ & Size (MB) $\downarrow$ & Lat. (ms) $\downarrow$ & Speedup $\uparrow$
& Acc. (\%) $\uparrow$ & Size (MB) $\downarrow$ & Lat. (ms) $\downarrow$ & Speedup $\uparrow$
& Acc. (\%) $\uparrow$ & Size (MB) $\downarrow$ & Lat. (ms) $\downarrow$ & Speedup $\uparrow$ \\
\midrule
\multicolumn{13}{c}{\textbf{(a) Main results}} \\
\midrule

Baseline
& 78.37 & 42.65 & 2.45 & 1.00
& 76.03 & 139.38 & 3.42 & 1.00
& 79.38 & 56.18 & 2.62 & 1.00 \\

Prune (30\%)
& \underline{79.84} & 33.13 & 2.43 & 1.01
& \textbf{79.41} & 97.56 & 3.43 & 1.00
& 80.11 & 39.33 & 2.64 & 0.99 \\

Prune (50\%)
& \textbf{80.38} & 26.97 & 2.55 & 0.96
& \underline{79.07} & 69.69 & 3.36 & 1.02
& \underline{80.55} & 28.11 & 2.81 & 0.93 \\

QAT (INT8)
& 77.42 & \underline{10.66} & \textbf{0.99} & \textbf{2.47}
& 71.29 & \underline{35.81} & \textbf{1.42} & \textbf{2.41}
& 77.23 & \underline{14.05} & \underline{1.01} & \underline{2.59} \\

KD
& 76.33 & 42.65 & 2.49 & 0.98
& 72.06 & 139.38 & 3.34 & 1.02
& 78.57 & 56.18 & 2.69 & 0.97 \\

Hybrid (Prune50\% $\rightarrow$ QAT $\rightarrow$ KD)
& 79.62 & \textbf{6.74} & \underline{1.00} & \underline{2.45}
& 78.69 & \textbf{17.41} & \underline{1.42} & \underline{2.41}
& \textbf{81.42} & \textbf{7.03} & \textbf{0.99} & \textbf{2.65} \\

\midrule
\multicolumn{13}{c}{\textbf{(b) Stage-order ablation (same components, same 20/40/40 budgets)}} \\
\midrule

Hybrid (Prune50\% $\rightarrow$ QAT $\rightarrow$ KD) \textit{(default)}
& \textbf{79.62} & \textbf{6.74} & \textbf{1.00} & \textbf{2.45}
& \textbf{78.69} & \textbf{17.41} & \underline{1.42} & \underline{2.41}
& \textbf{81.42} & \textbf{7.03} & \underline{0.99} & \underline{2.65} \\

Hybrid (Prune50\% $\rightarrow$ KD $\rightarrow$ QAT)
& \underline{79.00} & 6.74 & 1.02 & 2.40
& \underline{78.20} & 17.41 & \textbf{1.41} & \textbf{2.43}
& \underline{80.90} & 7.03 & \textbf{0.98} & \textbf{2.67} \\

Hybrid (QAT $\rightarrow$ Prune50\% $\rightarrow$ KD)
& 78.50 & 6.74 & 1.00 & 2.45
& 77.40 & 17.41 & 1.43 & 2.39
& 80.30 & 7.03 & 0.99 & 2.65 \\

Hybrid (QAT $\rightarrow$ KD $\rightarrow$ Prune50\%)
& 76.60 & 6.74 & \underline{1.01} & \underline{2.43}
& 75.10 & 17.41 & 1.42 & 2.41
& 78.80 & 7.03 & 1.00 & 2.62 \\
\bottomrule
\end{tabular}}
\vspace{2pt}
{\scriptsize \textit{Legend:} \textbf{Bold}/\underline{underline} indicate the best/second-best entries within each backbone block, computed separately for (a) and (b); ties are broken in favor of the \emph{default} ordering in (b).}

{\scriptsize \textit{Notes:} All hybrid variants use the same stage budgets (20/40/40 for Prune/QAT/KD). Speedup is relative to the FP32 baseline; minor differences may arise from rounding.}

\end{table*}

\subsection{Main Results: Accuracy--Size--Latency Trade-off}
\label{subsec:exp_multimodel}

Table~\ref{tab:tradeoff_multisetting} reports our main results under the fully trained protocol (fixed 100-epoch budget) on three backbone--dataset pairs.
Across all settings, the ordered hybrid pipeline consistently achieves a favorable accuracy--size--latency trade-off compared with any single-stage baseline (as shown in Fig.~\ref{fig:synergy}).
In particular, pruning alone tends to reduce the serialized footprint but yields limited CPU speedup, whereas INT8 QAT provides the most reliable latency reduction; combining pruning, INT8 QAT, and KD in the proposed order improves the final accuracy at essentially the same sparse-INT8 deployment form.

All methods are evaluated at a consistent deployable endpoint (sparse INT8) and under a fixed training budget, so the observed differences mainly reflect optimization outcomes rather than implementation choices.
Overall, the gains are stable across both residual (ResNet/WRN) and VGG-style architectures, supporting the generality of the proposed recipe.

\noindent\textbf{Controlled ordering ablation.}
To isolate the effect of stage sequencing, we keep the same ingredients (pruning, INT8 QAT, and KD) and the same per-stage budgets (20/40/40), and only permute the stage order (Table~\ref{tab:tradeoff_multisetting}).
The default ordering (Prune$\rightarrow$QAT$\rightarrow$KD) consistently yields the best accuracy across backbones, while moving pruning to the end (QAT$\rightarrow$KD$\rightarrow$Prune) causes the largest degradation.
Notably, latency remains within a narrow range under the same sparse-INT8 deployment setting, so the accuracy gaps directly reflect the impact of ordering.
We next provide a diagnostic analysis to further explain why this ordering improves optimization in the sparse INT8 regime.

\subsection{Why Ordering Matters: Diagnostic Analysis on ResNet-18/CIFAR-10}
\label{subsec:exp_resnet18_diag}

\begin{figure}[t]
    \centering
    \includegraphics[width=0.8\columnwidth]{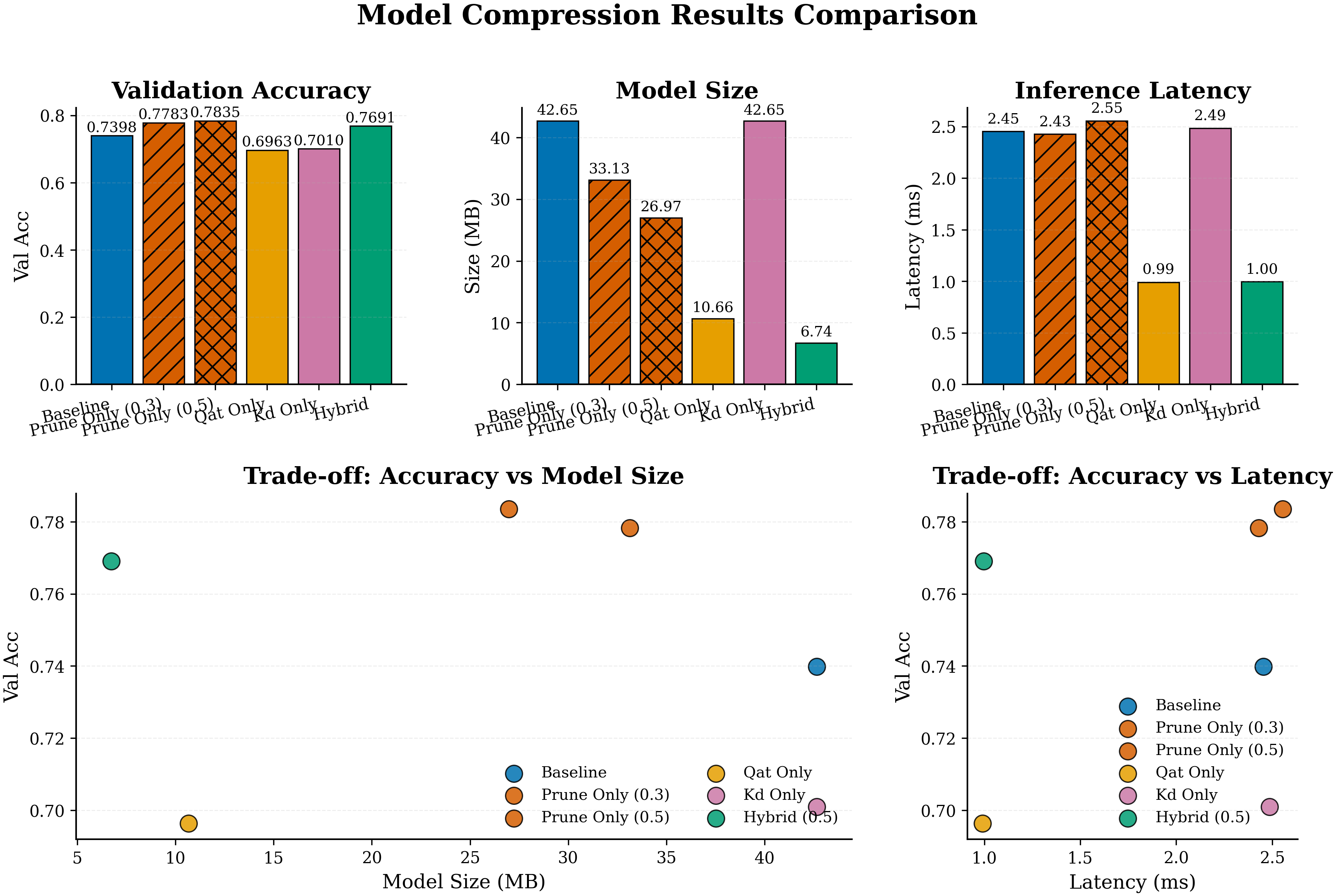} 
    \caption{Comparison of standalone baselines and the ordered hybrid pipeline on ResNet-18/CIFAR-10.}
    \label{fig:compare_results}
\end{figure}

We use the diagnostic snapshot on ResNet-18/CIFAR-10 (Table~\ref{tab:tradeoff_resnet18}) to analyze stage interactions and ordering effects, rather than to claim final performance.
Fig.~\ref{fig:compare_results} visualizes the resulting accuracy--efficiency trade-offs.

A key hardware observation is that \textbf{unstructured pruning does not necessarily translate to wall-clock speedups on standard CPUs}.
For example, 50\% pruning reduces the serialized footprint (42.65$\rightarrow$26.97\,MB) but slightly increases latency (2.45$\rightarrow$2.55\,ms), indicating that pruning mainly acts as capacity reduction and a pre-conditioner for subsequent optimization.
In contrast, \textbf{INT8 QAT provides the dominant latency reduction} (0.99\,ms, 2.48$\times$ speedup), yet it is harder to optimize in the early snapshot and suffers the largest accuracy drop (Val.\ 73.98$\rightarrow$69.63).
These observations motivate separating the roles of stages: pruning for capacity/conditioning, QAT for latency, and KD for accuracy recovery.

\begin{figure*}[!t]
    \centering
    \begin{subfigure}[t]{0.32\textwidth}
        \centering
        \includegraphics[width=\linewidth,trim=0pt 5pt 0pt 0pt,clip]{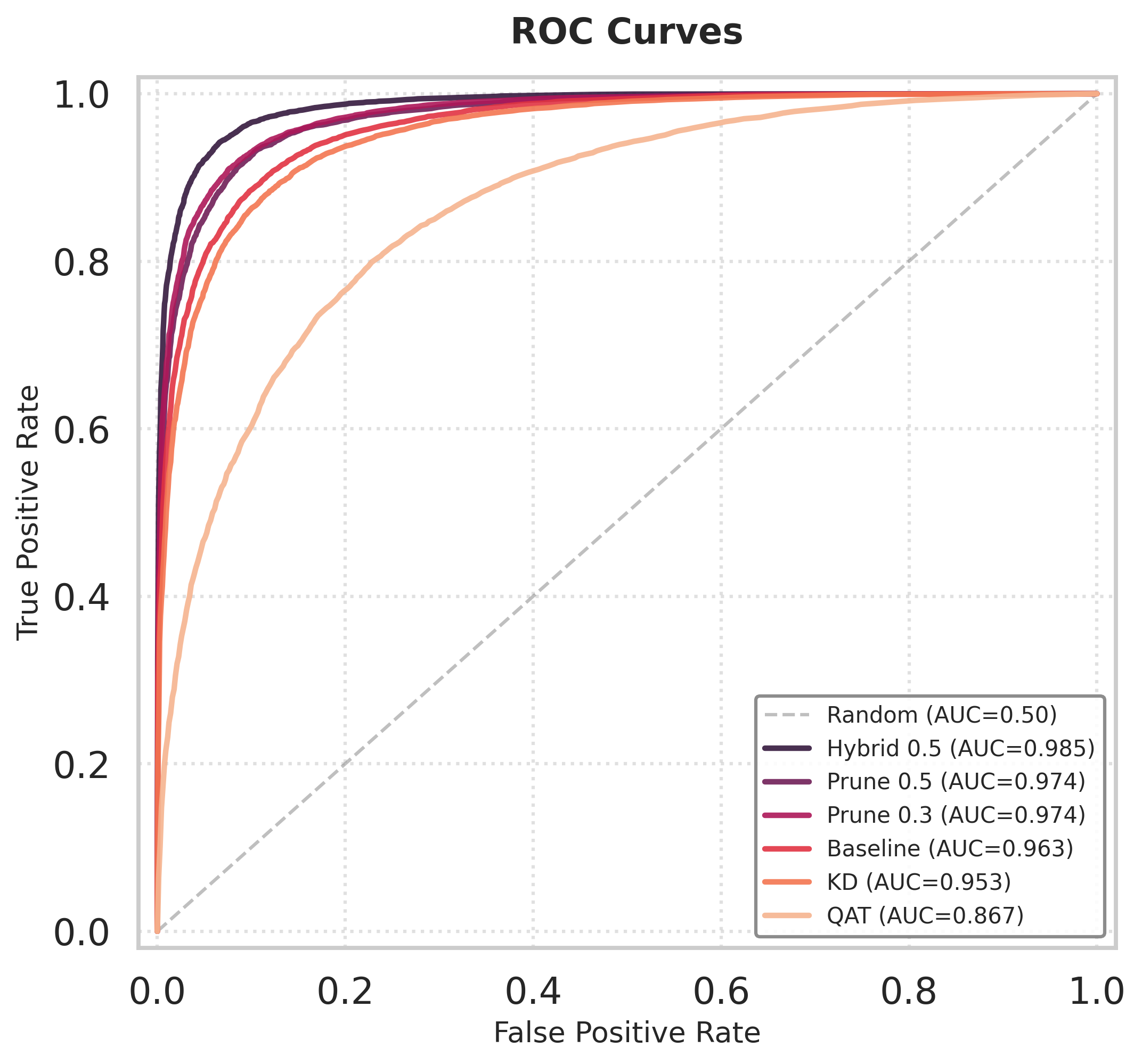}
        \caption{ROC curves.}
        \label{fig:roc}
    \end{subfigure}\hfill
    \begin{subfigure}[t]{0.32\textwidth}
        \centering
        \includegraphics[width=\linewidth,trim=0pt 5pt 0pt 0pt,clip]{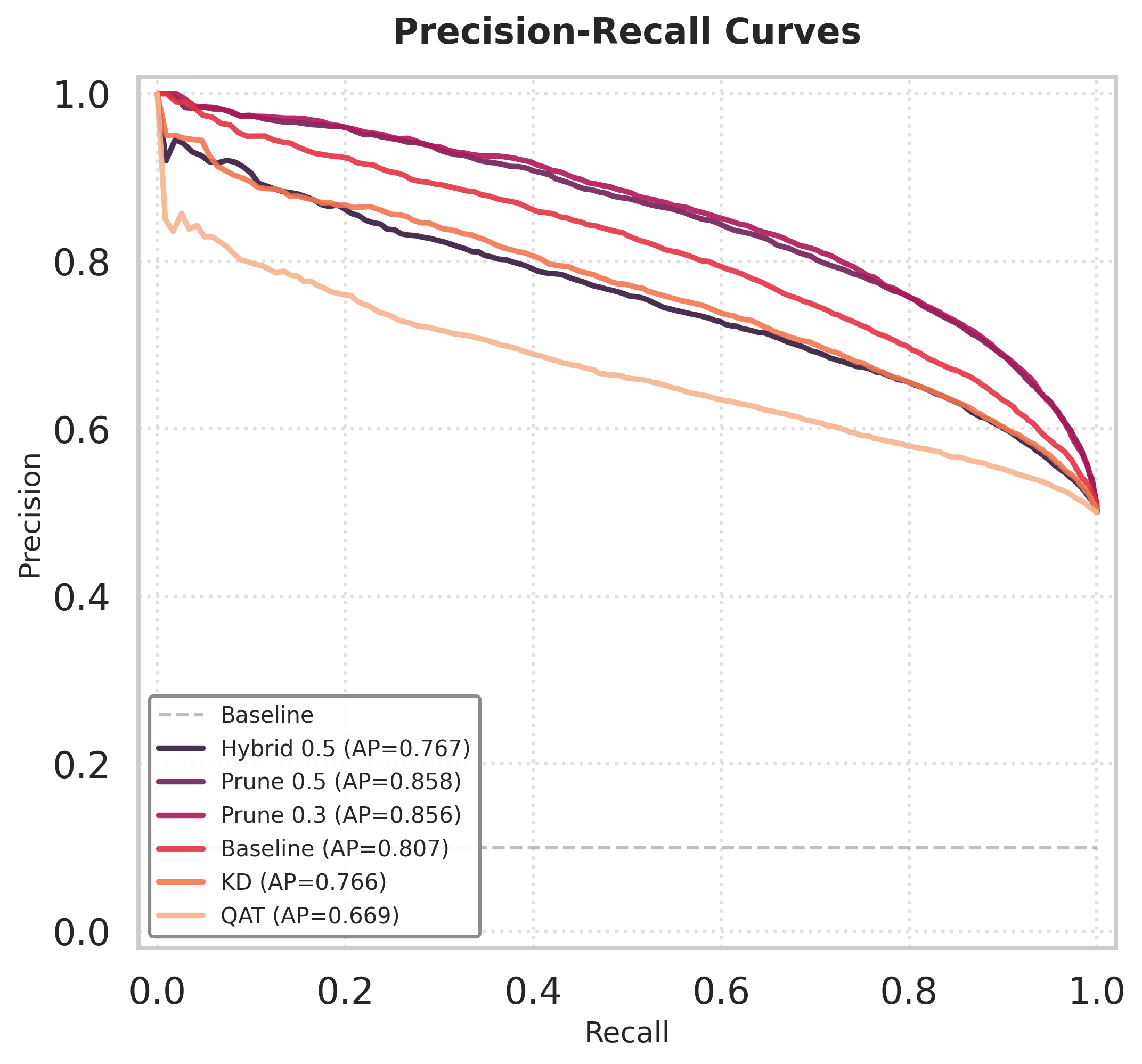}
        \caption{Precision--recall curves.}
        \label{fig:pr}
    \end{subfigure}\hfill
    \begin{subfigure}[t]{0.32\textwidth}
        \centering
        \includegraphics[width=\linewidth,trim=15pt 10pt 20pt 5pt,clip]{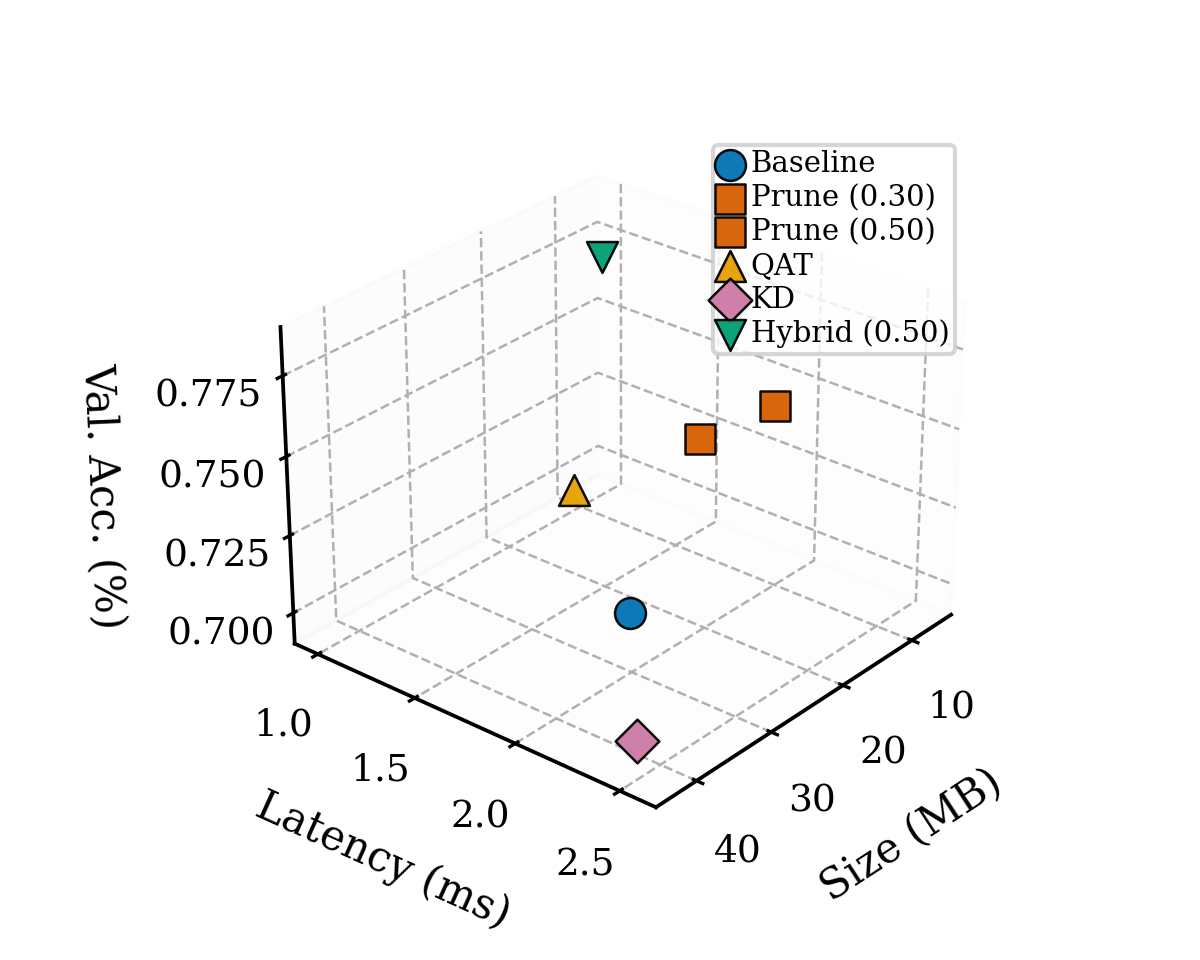}
        \caption{3D accuracy--size--latency trade-off.}
        \label{fig:3d}
    \end{subfigure}
    \caption{Global analysis on ResNet-18/CIFAR-10: ROC/PR curves and the 3D accuracy--size--latency trade-off.}
    \label{fig:global_analysis}
\end{figure*}

The ordered pipeline resolves this tension.
Compared with QAT-only, the \emph{Hybrid} pipeline preserves essentially the same latency (1.00 vs.\ 0.99\,ms) while recovering accuracy to 76.91\% (+7.28 points), yielding a substantially stronger accuracy--size--latency trade-off.
Moreover, it achieves the highest compression ratio among all compared variants (6.33$\times$), highlighting that the gains come with a compact deployable model.
The 3D view in Fig.~\ref{fig:3d} further illustrates how the proposed ordering shifts the solution toward a more favorable region in the joint trade-off space.
As auxiliary diagnostics, ROC/PR curves in Fig.~\ref{fig:global_analysis} show consistent trends with the main accuracy--efficiency comparisons.

\begin{table}[t]
    \centering
    \caption{Diagnostic snapshot on ResNet-18/CIFAR-10.}
    \label{tab:tradeoff_resnet18}
    \scriptsize
    \setlength{\tabcolsep}{1.6pt}
    \renewcommand{\arraystretch}{1.08}

    \resizebox{\columnwidth}{!}{
    \begin{tabular}{lcc
                    S[table-format=2.2]
                    S[table-format=2.2]
                    S[table-format=2.2]
                    S[table-format=1.2]
                    S[table-format=1.2]
                    S[table-format=1.2]}
        \toprule
        \multirow{2}{*}{\textbf{Method}} 
        & \multirow{2}{*}{\textbf{Ratio}} 
        & \multirow{2}{*}{\textbf{FT}} 
        & \multicolumn{2}{c}{\textbf{Acc. (\%) $\uparrow$}} 
        & \multicolumn{4}{c}{\textbf{Efficiency}} \\
        \cmidrule(lr){4-5} \cmidrule(lr){6-9}
        & & 
        & {Tr.} & {Val.} 
        & {Size (MB) $\downarrow$} 
        & {Compr. ($\times$) $\uparrow$} 
        & {Lat. (ms) $\downarrow$} 
        & {Spdup ($\times$) $\uparrow$} \\
        \midrule
        Baseline   & --   & 20 & 75.51 & 73.98 & 42.65 & 1.00 & 2.45 & 1.00 \\
        Prune Only & 30\% & 10 & \underline{81.75} & \underline{77.83} & 33.13 & 1.29 & \underline{2.43} & \underline{1.01} \\
        Prune Only & 50\% & 10 & \textbf{81.77} & \textbf{78.35} & 26.97 & 1.58 & 2.55 & 0.96 \\
        QAT Only   & --   &  8 & 72.42 & 69.63 & \underline{10.66} & \underline{4.00} & \textbf{0.99} & \textbf{2.48} \\
        KD Only    & --   & 10 & 70.41 & 70.10 & 42.65 & 1.00 & 2.49 & 0.99 \\
        \midrule
        Hybrid     & 50\% & 28 & 80.67 & 76.91 & \textbf{6.74} & \textbf{6.33}$^{\star}$ & \underline{1.00} & \underline{2.46} \\
        \bottomrule
    \end{tabular}
    }

    \vspace{2pt}
    {\scriptsize \textit{Notes:} \textbf{FT} = fine-tuning epochs; \textit{Size} = checkpoint MB. $^{\star}$ best compression.}
\end{table}

\begin{table}[t]
\centering
\caption{Literature-aligned comparison on ResNet-20/CIFAR-10.}
\label{tab:cifar10_rn20_300e_bops}
\footnotesize
\setlength{\tabcolsep}{2.8pt}
\renewcommand{\arraystretch}{1.05}
\begin{tabular}{lcccc}
\toprule
\textbf{Method} & \textbf{W-bit} & \textbf{Prune} & \textbf{Acc. (\%)} $\uparrow$ & \textbf{Rel. BOPs} $\downarrow$ \\
\midrule
Baseline & 32 & None  & 91.70 & 100.0 \\
SQL     & MP & Unstr. & 90.90 & 6.1 \\
QST-P   & MP & Unstr. & 91.80 & 5.0 \\
GETA    & MP & Str.   & 91.42 & 4.5 \\
QST-Q   & MP & Unstr. & 91.60 & 3.3 \\
\textbf{Ours (P50$\rightarrow$INT8QAT$\rightarrow$KD)} & \textbf{8} & Unstr. & \textbf{91.83} & \textbf{3.1} \\
\bottomrule
\end{tabular}

\vspace{2pt}
{\scriptsize \textit{Notes:} MP denotes mixed-precision. Rel.\ BOPs follow each paper's convention; ours uses W8/A8 INT8 QAT with 50\% unstructured sparsity.}
\end{table}

\subsection{Literature-aligned Comparison on ResNet-20/CIFAR-10}
\label{subsec:exp_lit_aligned}

Following the standard training protocol used in prior co-compression studies, we report a literature-aligned comparison on ResNet-20/CIFAR-10 in Table~\ref{tab:cifar10_rn20_300e_bops}, using accuracy and relative BOPs as commonly adopted in this line of work.

As shown in Table~\ref{tab:cifar10_rn20_300e_bops}, our ordered pipeline attains strong accuracy (91.83\%) while achieving the lowest relative BOPs (3.1) among the compared methods.
Notably, this trade-off is obtained with a simple and deployable recipe (50\% unstructured pruning with W8/A8 INT8 QAT and final-stage KD), rather than a heavily coupled mixed-precision objective.
Together with the deployment-driven CPU latency results in Table~\ref{tab:tradeoff_multisetting}, these findings suggest that the proposed ordering remains competitive under both measured runtime and literature-aligned proxy metrics.

\section{Conclusion}
\label{sec:conclusion}

We study an ordered compression pipeline---Prune$\rightarrow$INT8 QAT$\rightarrow$KD---and show that its stages play complementary roles.
On standard CPUs, unstructured pruning alone does not guarantee wall-clock speedups, but it reduces capacity and stabilizes subsequent low-precision optimization by shrinking the active weight set.
INT8 QAT contributes most of the latency reduction, while KD, applied last, recovers accuracy within the already constrained sparse INT8 regime.
Under a fixed training budget and a consistent deployable endpoint (sparse INT8), we observe consistently favorable accuracy--size--latency trade-offs across ResNet-18, WRN-28-10, and VGG-16-BN on CIFAR-10/100 compared with single-stage baselines, and controlled ordering ablations further confirm that stage sequencing is consequential.
Overall, our results highlight a practical takeaway: deployment decisions should be guided by measured latency alongside accuracy, rather than parameter/FLOP counts alone.
We further show that the proposed ordering remains competitive under a literature-aligned proxy metric (relative BOPs) on ResNet-20/CIFAR-10.
Finally, the pipeline is modular and can incorporate alternative pruning criteria, quantization schemes, or distillation losses without changing the overall recipe.

Future work will explore hardware-friendly structured sparsity and more automated policy selection to better map Pareto frontiers under different deployment constraints.

\FloatBarrier
\bibliographystyle{IEEEtran}
\bibliography{IEEEabrv,References}

\end{document}